\documentclass[11pt,a4paper]{article}
\usepackage[hyperref]{naaclhlt2018}
\usepackage{times}
\usepackage{latexsym}
\usepackage{amsmath}
\usepackage{graphics}
\usepackage{graphicx} 

\usepackage{url}

\aclfinalcopy 
\def\aclpaperid{24} 

\title{UnibucKernel: A kernel-based learning method for complex word identification}

\author{Andrei M. Butnaru \and {Radu Tudor} Ionescu\\
  \\
  University of Bucharest\\
  Department of Computer Science\\
  14 Academiei, Bucharest, Romania\\
  {\tt butnaruandreimadalin@gmail.com}\\
  {\tt raducu.ionescu@gmail.com}
}

\date{}

\begin{document}
\maketitle
\begin{abstract}
In this paper, we present a kernel-based learning approach for the 2018 Complex Word Identification (CWI) Shared Task. Our approach is based on combining multiple low-level features, such as character n-grams, with high-level semantic features that are either automatically learned using word embeddings or extracted from a lexical knowledge base, namely WordNet. After feature extraction, we employ a kernel method for the learning phase. The feature matrix is first transformed into a normalized kernel matrix. For the binary classification task (simple versus complex), we employ Support Vector Machines. For the regression task, in which we have to predict the complexity level of a word (a word is more complex if it is labeled as complex by more annotators), we employ $\nu$-Support Vector Regression. We applied our approach only on the three English data sets containing documents from Wikipedia, WikiNews and News domains. Our best result during the competition was the third place on the English Wikipedia data set. However, in this paper, we also report better post-competition results.
\end{abstract}

\section{Introduction}
\label{intro}

A key role in reading comprehension by non-native speakers is played by lexical complexity. To date, researchers in the Natural Language Processing (NLP) community have developed several systems to simply texts for non-native speakers \cite{Petersen-SLATE-2007} as well as native speakers with reading disabilities \cite{Rello-W4A-2013} or low literacy levels \cite{Specia-PROPOR-2010}. The first task that needs to be addressed by text simplification methods is to identify which words are likely to be considered complex. The complex word identification (CWI) task raised a lot of attention in the NLP community, as it has been addressed as a stand-alone task by some researchers \cite{Davoodi-ICSC-2017}. More recently, researchers even organized shared tasks on CWI \cite{paetzold-specia:2016:SemEval1,stajner-EtAl:2018:BEA}. The goal of the 2018 CWI Shared Task \cite{stajner-EtAl:2018:BEA} is to predict which words can be difficult for a non-native speaker, based on annotations collected from a mixture of native and non-native speakers. Although the task features a multilingual data set, we participated only in the English monolingual track, due to time constraints. In this paper, we describe the approach of our team, UnibucKernel, for the English monolingual track of the 2018 CWI Shared Task \cite{stajner-EtAl:2018:BEA}. We present results for both classification (predicting if a word is simple or complex) and regression (predicting the complexity level of a word) tasks. Our approach is based on a standard machine learning pipeline that consists of two phases: $(i)$ feature extraction and $(ii)$ classification/regression. In the first phase, we combine multiple low-level features, such as character n-grams, with high-level semantic features that are either automatically learned using word embeddings \cite{Mikolov-NIPS-2013} or extracted from a lexical knowledge base, namely WordNet \cite{Miller-WN-1995,Fellbaum-WN-1998}. After feature extraction, we employ a kernel method for the learning phase. The feature matrix is first transformed into a normalized kernel matrix, using either the inner product between pairs of samples (computed by the linear kernel function) or an exponential transformation of the inner product (computed by the Gaussian kernel function). For the binary classification task, we employ Support Vector Machines (SVM) \cite{cortes-vapnik-ml-1995}, while for the regression task, we employ $\nu$-Support Vector Regression (SVR) \cite{Chang-NC-2002}. We applied our approach only on the three English monolingual data sets containing documents from Wikipedia, WikiNews and News domains. Our best result during the competition was the third place on the English Wikipedia data set. Nonetheless, in this paper, we also report better post-competition results.

The rest of this paper is organized as follows. Related work on complex word identification is presented in Section \ref{sec_RelatedWork}. Our method is presented in Section \ref{sec_Method}. Our experiments and results are presented in Section \ref{sec_Experiments}. Finally, we draw our conclusions and discuss future work in Section \ref{sec_Conclusion}.

\section{Related Work}
\label{sec_RelatedWork}

Although text simplification methods have been proposed since more than a couple of years ago \cite{Petersen-SLATE-2007}, complex word identification has not been studied as a stand-alone task until recently \cite{Shardlow-SRW-2013}, with the first shared task on CWI organized in 2016 \cite{paetzold-specia:2016:SemEval1}. With some exceptions \cite{Davoodi-ICSC-2017}, most of the related works are actually the system description papers of the 2016 CWI Shared Task participants. Among the top $10$ participants, the most popular classifier is Random Forest \cite{brooke-uitdenbogerd-baldwin:2016:SemEval,mukherjee-EtAl:2016:SemEval,ronzano-EtAl:2016:SemEval,zampieri-tan-vangenabith:2016:SemEval}, while the most common type of features are lexical and semantic features \cite{brooke-uitdenbogerd-baldwin:2016:SemEval,mukherjee-EtAl:2016:SemEval,paetzold-specia:2016:SemEval2,quijada-medero:2016:SemEval,ronzano-EtAl:2016:SemEval}. Some works used Naive Bayes \cite{mukherjee-EtAl:2016:SemEval} or SVM \cite{zampieri-tan-vangenabith:2016:SemEval} along with the Random Forest classifier, while others used different classification methods altogether, e.g. Decision Trees \cite{quijada-medero:2016:SemEval}, Nearest Centroid \cite{palakurthi-mamidi:2016:SemEval} or Maximum Entropy \cite{konkol:2016:SemEval}. Along with the lexical and semantic features, many have used morphological \cite{mukherjee-EtAl:2016:SemEval,paetzold-specia:2016:SemEval2,palakurthi-mamidi:2016:SemEval,ronzano-EtAl:2016:SemEval} and syntactic \cite{mukherjee-EtAl:2016:SemEval,quijada-medero:2016:SemEval,ronzano-EtAl:2016:SemEval} features.

\newcite{paetzold-specia:2016:SemEval2} proposed two ensemble methods by applying either hard voting or soft voting on machine learning classifiers trained on morphological, lexical, and semantic features. Their systems ranked on the first and the second places in the 2016 CWI Shared Task. \newcite{ronzano-EtAl:2016:SemEval} employed Random Forests based on lexical, morphological, semantic and syntactic features, ranking on the third place in the 2016 CWI Shared Task. \newcite{konkol:2016:SemEval} trained Maximum Entropy classifiers on word occurrence counts in Wikipedia documents, ranking on the fourth place, after \newcite{ronzano-EtAl:2016:SemEval}. \newcite{wrobel:2016:SemEval} ranked on fifth place using a simple rule-based approach that considers one feature, namely the number of documents from Simple English Wikipedia in which the target word occurs. \newcite{mukherjee-EtAl:2016:SemEval} employed the Random Forest and the Naive Bayes classifiers based on semantic, lexicon-based, morphological and syntactic features. Their Naive Bayes system ranked on the sixth place in the 2016 CWI Shared Task. After the 2016 CWI Shared Task, \newcite{zampieri-EtAl:2017:NLPTEA} combined the submitted systems using an ensemble method based on plurality voting. They also proposed an oracle ensemble that provides a theoretical upper bound of the performance. The oracle selects the correct label for a given word if at least one of the participants predicted the correct label. The results reported by \newcite{zampieri-EtAl:2017:NLPTEA} 
indicate that there is a significant performance gap to be filled by automatic systems.

Compared to the related works, we propose the use of some novel semantic features. One set of features is inspired by the work of \newcite{ShotgunWSD-EACL-2017} in word sense disambiguation, while another set of features is inspired by the spatial pyramid approach \cite{Lazebnik-BBF-2006}, commonly used in computer vision to improve the performance of the bag-of-visual-words model \cite{radu-WREPL-2013,Ionescu_Popescu_PRL_2014}.

\section{Method}
\label{sec_Method}

The method that we employ for identifying complex words is based on a series of features extracted from the word itself as well as the context in which the word is used. Upon having the features extracted, we compute a kernel matrix using one of two standard kernel functions, namely the linear kernel or the Gaussian kernel. We then apply either the SVM classifier to identify if a word is complex or not, or the $\nu$-SVR predictor to determine the complexity level of a word.

\subsection{Feature Extraction}

As stated before, we extract features from both the target word and the context in which the word appears. Form the target word, we quantify a series of features based on its characters. More specifically, we count the number of characters, vowels and constants, as well as the percentage of vowels and constants from the total number of characters in the word. Along with these features, we also quantify the number of consecutively repeating characters, e.g. double consonants. For example, in the word ``innovation'', we can find the double consonant ``nn''. We also extract n-grams of 1, 2, 3 and 4 characters, based on the intuition that some complex words tend to be formed of a different set of n-grams than simple words. For instance, the complex word ``cognizant'' is formed of rare 3-grams, e.g. ``ogn'' or ``niz'', compared to its commonly-used synonym ``aware'', which contains 3-grams that we can easily find in other simple words, e.g. ``war'' or ``are''.

Other features extracted from the target word are the part-of-speech and the number of senses listed in the WordNet knowledge base \cite{Miller-WN-1995,Fellbaum-WN-1998}, for the respective word. If the complex word is actually composed of multiple words, i.e. it is a \emph{multi-word expression}, we generate the features for each word in the target and sum the corresponding values to obtain the features for the target multi-word expression.

In the NLP community, word embeddings \cite{Bengio-JMLR-2003,Collobert-ICML-2008} are used in many tasks, and became more popular due to the \emph{word2vec} \cite{Mikolov-NIPS-2013} framework. Word embeddings methods have the capacity to build a vectorial representation of words by assigning a low-dimensional real-valued vector to each word, with the property that semantically related words are projected in the same vicinity of the generated space. Word embeddings are in fact a learned representation of words where each dimension represents a hidden feature of the word \cite{Turian-ACL-2010}. We devise additional features for the CWI task with the help of pre-trained word embeddings provided by \emph{word2vec} \cite{Mikolov-NIPS-2013}. The first set of features based on word embeddings takes into account the word's context. More precisely, we record the minimum, the maximum and the mean value of the cosine similarity between the target word and each other word from the sentence in which the target word occurs. The intuition for using this set of features is that a word can be complex if it is semantically different from the other context words, and this difference should be reflected in the embedding space. Having the same goal in mind, namely to identify if the target word is an outlier with respect to the other words in the sentence, we employ a simple approach to compute sense embeddings using the semantic relations between WordNet synsets. We note that this approach was previously used for unsupervised word sense disambiguation in \cite{ShotgunWSD-EACL-2017}. To compute the sense embedding for a word sense, we first build a \emph{disambiguation vocabulary} or \emph{sense bag}. Based on WordNet, we form the sense bag for a given synset by collecting the words found in the gloss of the synset (examples included) as well as the words found in the glosses of semantically related synsets. The semantic relations are chosen based on the part-of-speech of the target word, as described in \cite{ShotgunWSD-EACL-2017}. To derive the sense embedding, we embed the collected words in an embedding space and compute the median of the resulted word vectors. For each sense embedding of the target word, we compute the cosine similarity with each and every sense embedding computed for each other word in the sentence, in order to find the minimum, the maximum and the mean value. 


\begin{figure}[t]
\centering
\includegraphics[width=1\linewidth]{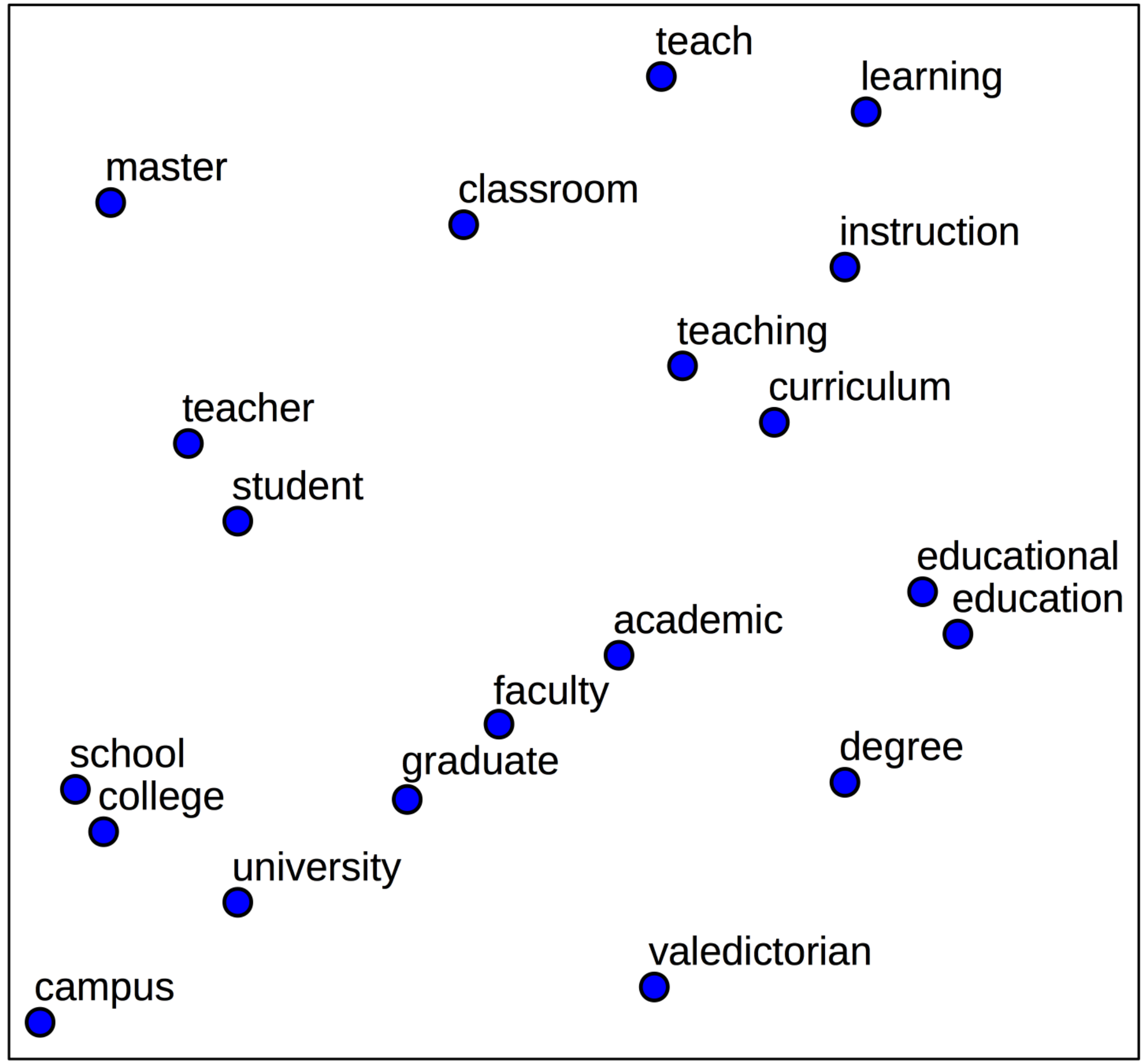}
\caption{A set of word vectors represented in a 2D space generated by applying PCA on 300-dimensional word embeddings.\label{Fig_WE_2_dim}}
\end{figure}

\begin{figure}[t]
\centering
\includegraphics[width=1\linewidth]{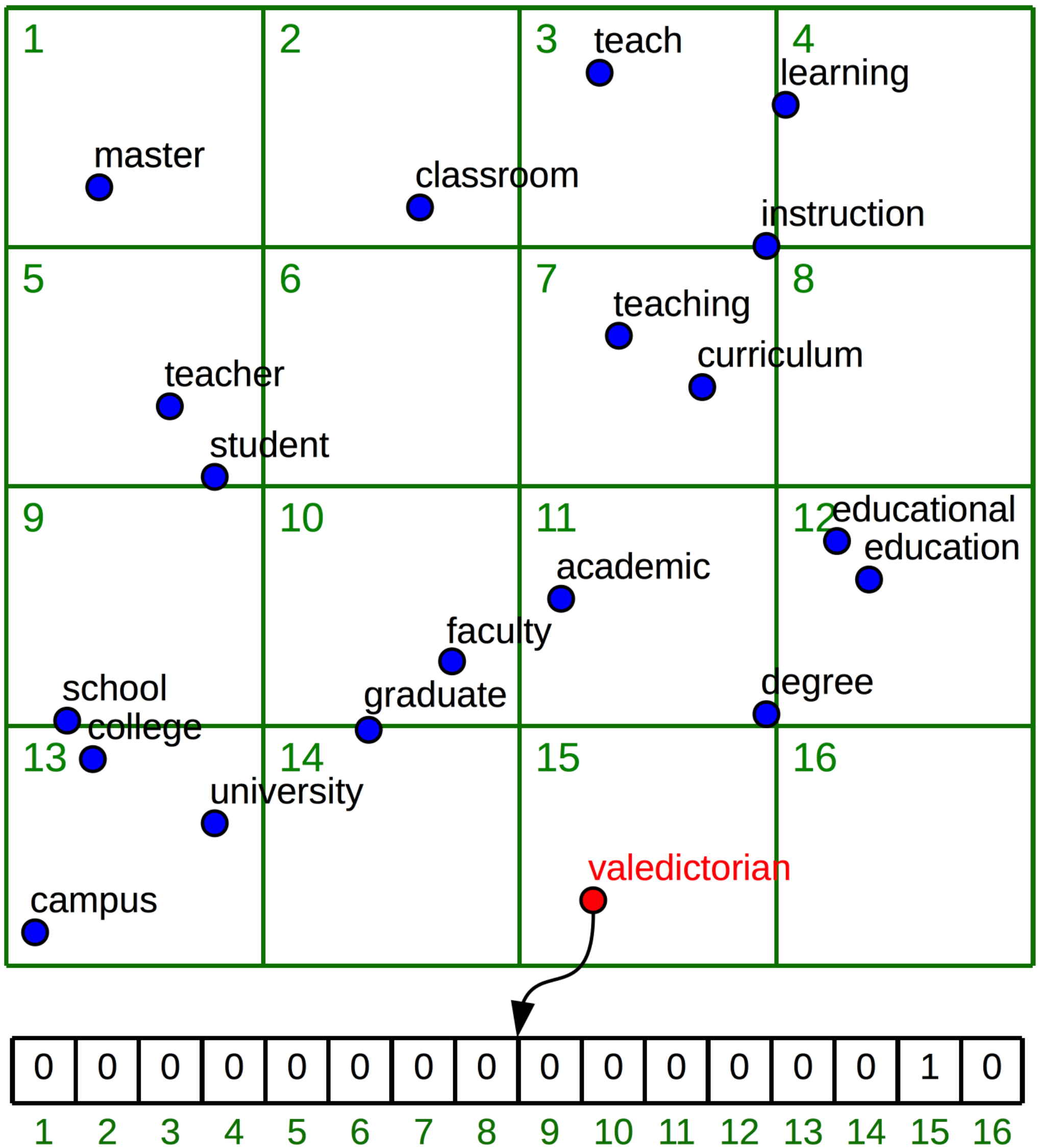}
\caption{A grid of $4 \times 4$ applied on the 2D embedding space. For example, the word ``valedictorian'' is located in bin number $15$. Consequently, the corresponding one-hot vector contains a non-zero value at index $15$. \label{Fig_WE_2_dim_bins}}
\end{figure}

Using pre-trained word embeddings provided by the \emph{GloVe} framework \cite{Pennington-EMNLP-2014}, we further managed to define a set of useful features based on the location of the target word in the embedding space. In this last set of features, we first process the word vectors in order to reduce the dimensionality of the vector space from 300 components to only 2 components, by applying Principal Component Analysis (PCA) \cite{hotelling-PCA-1933}. Figure \ref{Fig_WE_2_dim} illustrates a couple of semantically related words, that are projected in the same area of the 2-dimensional (2D) embedding space generated by PCA. In the newly generated space, we apply a grid to divide the space into multiple and equal regions, named bins. This process is inspired by the spatial pyramids \cite{Lazebnik-BBF-2006} used in computer vision to recover spatial information in the bag-of-visual-words \cite{radu-WREPL-2013,Ionescu_Popescu_PRL_2014}. After we determine the bins, we index the bins and encode the index of the bin that contains the target word as a one-hot vector. Various grid sizes could provide a more specific or a more general location of a word in the generated space. For this reason, we use multiple grid sizes starting from coarse divisions such as $2 \times 2$, $4 \times 4$, and $8 \times 8$, to fine divisions such as $16 \times 16$ and $32 \times 32$. In Figure \ref{Fig_WE_2_dim_bins}, we show an example with a $4 \times 4$ grid that divides the space illustrated in Figure\ref{Fig_WE_2_dim} into $16$ bins, and the word ``valedictorian'' is found in bin number $15$. The corresponding one-hot vector, containing a single non-zero value at index $15$, is also illustrated in Figure \ref{Fig_WE_2_dim_bins}. The thought process for using this one-hot representation is that complex words tend to reside alone in the semantic space generated by the word embedding framework.

We would like to point out that each and every type of features described in this section has a positive influence on the overall accuracy of our framework.

\subsection{Kernel Representation}

Kernel-based learning algorithms work by embedding the data into a Hilbert space and by searching for linear relations in that space, using a learning algorithm. The embedding is performed implicitly, that is by specifying the inner product between each pair of points rather than by giving their coordinates explicitly. The power of kernel methods \cite{radu-marius-book-2016,taylor-Cristianini-cup-2004} lies in the implicit use of a Reproducing Kernel Hilbert Space induced by a positive semi-definite kernel function. Despite the fact that the mathematical meaning of a kernel is the inner product in a Hilbert space, another interpretation of a kernel is the pairwise similarity between samples.

The kernel function offers to the kernel methods the power to naturally handle input data that is not in the form of numerical vectors, such as strings, images, or even video and audio files. The kernel function captures the intuitive notion of similarity between objects in a specific domain and can be any function defined on the respective domain that is symmetric and positive definite. In our approach, we experiment with two commonly-used kernel functions, namely the linear kernel and the Radial Basis Function (RBF) kernel. The \emph{linear kernel} is easily obtained by computing the inner product of two feature vectors $x$ and $z$:
\begin{equation*}
k(x,z) = \langle x, z \rangle ,
\end{equation*}
where $\langle \cdot, \cdot \rangle$ denotes the inner product. In a similar manner, the \emph{RBF kernel} (also known as the Gaussian kernel) between two feature vectors $x$ and $z$ can be computed as follows:
\begin{equation*}
k(x,z) = exp \left(- \frac{1 - \langle x, z \rangle}{2 \sigma^2} \right). 
\end{equation*}
In the experiments, we replace $1/{(2 \sigma^2)}$ with a constant value $r$, and tune the parameter $r$ instead of $\sigma$.

A technique that improves machine learning performance for many applications is data normalization. Because the range of raw data can have significant variation, the objective functions optimized by the classifiers will not work properly without normalization. The normalization step gives to each feature an approximately equal contribution to the similarity between two samples. The normalization of a pairwise kernel matrix $K$ containing similarities between samples is obtained by dividing each component to the square root of the product of the two corresponding diagonal elements: 

\begin{equation*}
	\hat{K}_{ij} = \dfrac{K_{ij}}{\sqrt{K_{ii} \cdot K_{jj}}}.
\end{equation*}

\subsection{Classification and Regression}

In the case of binary classification problems, kernel-based learning algorithms look for a discriminant function, a function that assigns $+1$ to examples that belong to one class and $-1$ to examples that belong to the other class. This function will be a linear function in the Hilbert space, which means it will have the form:
\begin{equation*}
\begin{split}
f(x)=\mbox{sign}(\langle w, x \rangle + b),
\end{split}
\end{equation*}
for some weight vector $w$ and some bias term $b$. The kernel can be employed whenever the weight vector can be expressed as a linear combination of the training points, $\sum\limits_{i=1}^n\alpha_i\: x_i$, implying that $f$ can be expressed as follows:
\begin{equation*}
\begin{split}
f(x)=\mbox{sign} \left( \sum\limits_{i=1}^n\alpha_i\: k(x_i,x)+b \right),
\end{split}
\end{equation*}
where $n$ is the number of training samples and $k$ is a kernel function.

Various kernel methods differ in the way they find the vector $w$ (or equivalently the dual vector $\alpha$). Support Vector Machines \cite{cortes-vapnik-ml-1995} try to find the vector $w$ that defines the hyperplane that maximally separates the images (outcomes of the embedding map) in the Hilbert space of the training examples belonging to the two classes. Mathematically, the SVM classifier chooses the weights $w$ and the bias term $b$ that satisfy the following optimization criterion:
\begin{equation*}
\begin{split}
\min_{w,b}\frac{1}{n}\sum\limits_{i=1}^n[1-y_i(\langle w,\phi(x_i) \rangle + b)]_+ + \nu ||w||^2 ,
\end{split}
\end{equation*}
where $y_i$ is the label ($+1$/$-1$) of the training example $x_i$, $\nu$ is a regularization parameter and $[x]_+=\max \lbrace x, 0 \rbrace$. We use the SVM classifier for the binary classification of words into simple versus complex classes. On the other hand, we employ $\nu$-Support Vector Regression ($\nu$-SVR) in order to predict the complexity level of a word (a word is more complex if it is labeled as complex by more annotators). The $\nu$-Support Vector Machines \cite{Chang-NC-2002} can handle both classification and regression. The model introduces a new parameter $\nu$, that can be used to control the amount of support vectors in the resulting model. The parameter $\nu$ is introduced directly into the optimization problem formulation and it is estimated automatically during training. 

\section{Experiments}
\label{sec_Experiments}

\subsection{Data Sets}

\begin{table}[t]
\small{
\begin{center}
\begin{tabular}{|l|r|r|r|}
\hline
Data Set 			& Train		    & Validation	& Test\\
\hline
\hline
English News		& $14002$		& $1764$		& $2095$\\
English WikiNews	& $7746$		& $870$		    & $1287$\\
English Wikipedia	& $5551$		& $694$			& $870$\\
\hline
\end{tabular}
\end{center}
}
\caption{\label{tab_Datasets} A summary with the number of samples in each data set of the English monolingual track of the 2018 CWI Shared Task.}
\end{table}

The data sets used in the English monolingual track of the 2018 CWI Shared Task \cite{stajner-EtAl:2018:BEA} are described in \cite{yimam-EtAl:2017:I17-2}. Each data set covers one of three distinct genres (News, WikiNews and Wikipedia), and the samples are annotated by both native and non-native English speakers. Table \ref{tab_Datasets} presents the number of samples in the training, the validation (development) and the test sets, for each of the three genres.

\subsection{Classification Results}

\begin{table*}[!t]
\small{
\begin{center}
\begin{tabular}{|l|l|c|c|c|c|c|c|}
\hline
Data Set 			& Kernel    & Accuracy	        & \multicolumn{2}{|c|}{$F_1$-score}   & Competition Rank    & Post-Competition Rank\\
\hline
\hline
English News		& linear	& $0.8653$	        & $0.8547$          & $0.8111^{*}$      & $12$              & $6$\\
English News		& RBF		& $0.8678$	        & $0.8594$          & $0.8178^{*}$      & $10$              & $5$\\
\hline
English WikiNews	& linear	& $0.8205$	        & $0.8151$          & $0.7786^{*}$      & $10$              & $5$\\
English WikiNews	& RBF	    & $0.8252$	        & $0.8201$          & $0.8127^{*}$      & $5$               & $4$\\
\hline
English Wikipedia	& linear	& $0.7874$	        & $0.7873$          & $0.7804^{*}$      & $6$               & $4$\\
English Wikipedia	& RBF	    & \;\;$0.7920^{*}$	& \;\;$0.7919^{*}$  & $0.7919^{*}$      & $3$               & $3$\\
\hline
\end{tabular}
\end{center}
}
\caption{\label{tab_CWI_classification} Classification results on the three data sets of the English monolingual track of the 2018 CWI Shared Task. The methods are evaluated in terms of the classification accuracy and the $F_1$-score. The results marked with an asterisk are obtained during the competition. The other results are obtained after the competition.}
\end{table*}

\noindent
{\bf Parameter Tuning.}
For the classification task, we used the SVM implementation provided by LibSVM \cite{LibSVM-2011}. The parameters that require tuning are the parameter $r$ of the RBF kernel and the regularization parameter $C$ of the SVM. We tune these parameters using grid search on each of the three validation sets included in the data sets prepared for the English monolingual track. For the parameter $r$, we select values from the set $\{ 0.5, 1.0, 1.5, 2.0 \}$. For the regularization parameter $C$ we choose values from the set $\{10^{-1}, 10^0, 10^1, 10^2 \}$. Interestingly, we obtain the best results with the same parameter choices on all three validation sets. The optimal parameter choices are $C = 10^1$ and $r = 1.0$. We use these parameters in all our subsequent classification experiments.

\begin{table*}[!t]
\small{
\begin{center}
\begin{tabular}{|l|l|c|c|}
\hline
Data Set 			& Kernel    & Mean Absolute Error   & Post-Competition Rank\\
\hline
\hline
English News		& linear	& $0.0573$              & $4$\\
English News		& RBF		& $0.0492$              & $1$\\
\hline
English WikiNews	& linear	& $0.0724$              & $4$\\
English WikiNews	& RBF	    & $0.0667$              & $1$\\
\hline
English Wikipedia	& linear	& $0.0846$              & $4$\\
English Wikipedia	& RBF	    & $0.0805$              & $2$\\
\hline
\end{tabular}
\end{center}
}
\caption{\label{tab_CWI_regression} Regression results on the three data sets of the English monolingual track of the 2018 CWI Shared Task. The methods are evaluated in terms of the mean absolute error (MAE). The reported results are obtained after the competition.}
\end{table*}

\noindent
{\bf Results.}
Our results for the classification task on the three data sets included in the English monolingual track are presented in Table \ref{tab_CWI_classification}. We would like to note that, before the competition ended, we observed a bug in the code that was used in most of our submissions. In the feature extraction stage, the code produced NaN (not a number) values for some features. In order to make the submissions in time, we had to eliminate the samples containing NaN values in the feature vector. Consequently, most of our results during the competition were lower than expected. However, we managed to fix this bug and recompute the features in time to resubmit new results, but only for the RBF kernel on the English Wikipedia data set. The rest of the results presented in Table \ref{tab_CWI_classification} are produced after the bug fixing and after the submission deadline. Nevertheless, for a fair comparison with the other systems, we include our $F_1$-scores and rankings during the competition as well as the post-competition $F_1$-scores and rankings.

The results reported in Table \ref{tab_CWI_classification} indicate that the RBF kernel is more suitable for the CWI task than the linear kernel. Our best $F_1$-score on the English News data set is $0.8594$, which is nearly $1.4\%$ lower than the top scoring system, which attained $0.8736$ during the competition. On the English WikiNews data set, our best $F_1$-score ($0.8201$) is once again about $2\%$ lower than the top scoring system, which obtained $0.8400$ during the competition. On the English Wikipedia data set, our best $F_1$-score is $0.7919$. With this score, we ranked as the third team on the English Wikipedia data set. Two systems performed better on English Wikipedia, one that reached the top $F_1$-score of $0.8115$ and one that reached the second-best scored of $0.7965$. Overall, our system performed quite well, but it can surely benefit from the addition of more features.

\subsection{Regression Results}

Although we did not submit results for the regression task, we present post-competition regression results in this section.

\noindent
{\bf Parameter Tuning.}
For the regression task, the parameters that require tuning are the parameter $r$ of the RBF kernel and the $\nu$-SVR parameters $C$ and $\nu$. As in the classification task, we tune these parameters using grid search on the validation sets provided with the three data sets included in the English monolingual track. For the parameter $r$, we select values from the set $\{ 0.5, 1.0, 1.5, 2.0 \}$. For the regularization parameter $C$ we choose values from the set $\{10^{-1}, 10^0, 10^1, 10^2 \}$. The preliminary results on the validation sets indicate the best parameter choices for each data set. For the English News data set, we obtained the best validation results using $C = 10^1$ and $r = 1.5$. For the English WikiNews and English Wikipedia data sets, we obtained the best validation results using $C = 10^0$ and $r = 1.5$. For the parameter $\nu$, we leave the default value of $0.5$ provided by LibSVM \cite{LibSVM-2011}.

\noindent
{\bf Results.}
The regression results on the three data sets included in the English monolingual track are presented in Table \ref{tab_CWI_regression}. The systems are evaluated in terms of the mean absolute error (MAE). As in the classification task, we can observe that the RBF kernel provides generally better results than the linear kernel. On two data sets, English News and English WikiNews, we obtain better MAE values than all the systems that participated in the competition. Indeed, the best MAE on English News reported during the competition is $0.0510$, and we obtain a smaller MAE ($0.0492$) using the RBF kernel. Similarly, with a MAE of $0.0667$ for the RBF kernel, we surpass the top system on English WikiNews, which attained a MAE of $0.0674$ during the competition. On the third data set, English Wikipedia, we attain the second-best score ($0.0805$), after the top system, that obtained a MAE of $0.0739$ during the competition. Compared to the classification task, we report better post-competition rankings in the regression task. This could be explained by two factors. First of all, the number of participants in the regression task was considerably lower. Second of all, we believe that $\nu$-SVR is a very good regressor which is not commonly used, surpassing alternative regression methods in other tasks as well, e.g. image difficulty prediction \cite{Ionescu-CVPR-2016}. 

\section{Conclusion}
\label{sec_Conclusion}

In this paper, we described the system developed by our team, UnibucKernel, for the 2018 CWI Shared Task. The system is based on extracting lexical, syntactic and semantic features and on training a kernel method for the prediction (classification and regression) tasks. We participated only in the English monolingual track. Our best result during the competition was the third place on the English Wikipedia data set. In this paper, we also reported better post-competition results.

In this work, we treated each English data set independently, due to the memory constraints of our machine. Nevertheless, we believe that joining the training sets provided in the English News, the English WikiNews and the English Wikipedia data sets into a single and larger training set can provide better performance, as the model's generalization capacity could improve by learning from an extended set of samples. We leave this idea for future work. Another direction that could be explored in future work is the addition of more features, as our current feature set is definitely far from being exhaustive.


\bibliography{CWI}
\bibliographystyle{acl_natbib}

\end{document}